\documentclass[letterpaper]{article} 
\usepackage{aaai2026}  
\usepackage{times}  
\usepackage{helvet}  
\usepackage{courier}  
\usepackage[hyphens]{url}  
\usepackage{graphicx} 
\usepackage{natbib}  
\usepackage{caption} 
\usepackage{algorithm}
\usepackage{algorithmic}
\usepackage{enumitem}
\usepackage{multirow}
\usepackage{amsthm}
\usepackage{amsmath}
\usepackage{amsfonts}
\usepackage{color}
\usepackage{pifont}
\usepackage[switch]{lineno}

\usepackage[table,xcdraw]{xcolor}
\usepackage[utf8]{inputenc} 
\usepackage[T1]{fontenc}    
\usepackage{url}            
\usepackage{booktabs}       
\usepackage{amsfonts}       
\usepackage{nicefrac}       
\usepackage{microtype}      

\usepackage{newfloat}
\usepackage{listings}
\DeclareCaptionStyle{ruled}{labelfont=normalfont,labelsep=colon,strut=off} 
\floatstyle{ruled}
\newfloat{listing}{tb}{lst}{}
\floatname{listing}{Listing}
\title{Coarse-to-Fine Open-Set Graph Node Classification \\
        with Large Language Models}
\author{
    Xueqi Ma\textsuperscript{\rm 1}, Xingjun Ma\textsuperscript{\rm 2}, Sarah Monazam Erfani\textsuperscript{\rm 1}, Danilo Mandic\textsuperscript{\rm 3}, James Bailey\textsuperscript{\rm 1}
}
\affiliations{
    \textsuperscript{\rm 1}The University of Melbourne, Australia\\

    \textsuperscript{\rm 2} Fudan University, China\\
    \textsuperscript{\rm 3} Imperial College London, UK\\
}

\usepackage{bibentry}

\begin{document}

\maketitle


\begin{abstract}

Developing open-set classification methods capable of classifying in-distribution (ID) data while detecting out-of-distribution (OOD) samples is essential for deploying graph neural networks (GNNs) in open-world scenarios. Existing methods typically treat all OOD samples as a single class, despite real-world applications—especially high-stake settings like fraud detection and medical diagnosis—demanding deeper insights into OOD samples, including their probable labels. This raises a critical question: \emph{Can OOD detection be extended to OOD classification without true label information?} To answer this question, we introduce a Coarse-to-Fine open-set Classification (CFC) method that leverages large language models (LLMs) for graph datasets. CFC consists of three key components: 1) A coarse classifier that utilizes LLM prompts for OOD detection and outlier label generation; 2) A GNN-based fine classifier trained with OOD samples from coarse classifier for enhanced OOD detection and ID classification; and 3) Refined OOD classification achieved through LLM prompts and post-processed OOD labels. Unlike methods relying on synthetic or auxiliary OOD samples, CFC employs semantic OOD data-instances that are genuinely out-of-distribution based on their inherent meaning, thus improving interpretability and practical utility. 
CFC enhances OOD detection by 10\% compared to state-of-the-art approaches on graph and text domain, while
achieving up to 70\% accuracy in OOD classification on graph datasets. The code is available at https://github.com/sihuo-design/CFC.
\end{abstract}

\section{Introduction}

Graph neural networks (GNNs) have demonstrated excellent performance in closed-set scenarios, where the train and test datasets share the same distributions. However, in many real-world applications, models are deployed on data containing previously unseen classes. Traditional GNN methods \cite{gcn, graphsage, gat} typically classify all unlabeled nodes into known classes, failing to identify nodes that belong to unknown classes, which degrades overall model performance. Addressing this limitation requires the development of models that can accurately classify in-distribution (ID) samples from known classes while effectively rejecting out-of-distribution (OOD) samples from unknown classes.  This critical challenge is commonly referred to as the open-set classification problem.

Recent approaches \cite{hendrycks2016baseline,song2022learning} to open-set node classification problem \cite{wu2021openwgl} have employed thresholding methods, using the maximum output probability as a confidence score to distinguish OOD samples from ID ones. While intuitive, determining an optimal threshold \cite{yang2023graph} to separate unknown from known classes is both challenging and time-consuming \cite{perera2020generative}.
Another line of research redefines open-set classification as a closed-set problem by estimating the distribution of unknown classes and adjusting the network's confidence. For example, methods such as \cite{ge2017generative,neal2018open,perera2020generative,zhou2021learning} generate synthetic samples as OOD, while others, like \cite{hendrycks2018deep, wang2023learning}, incorporate auxiliary training data—referred to as outlier exposure—to train image classifiers for OOD detection. Building on these works, \citet{zhang2023g2pxy} proposed generating proxy unknown nodes to simulate open-set data for graph node OOD detection. 

\begin{figure}[]  
    \centering
    \includegraphics[width=0.4\textwidth]{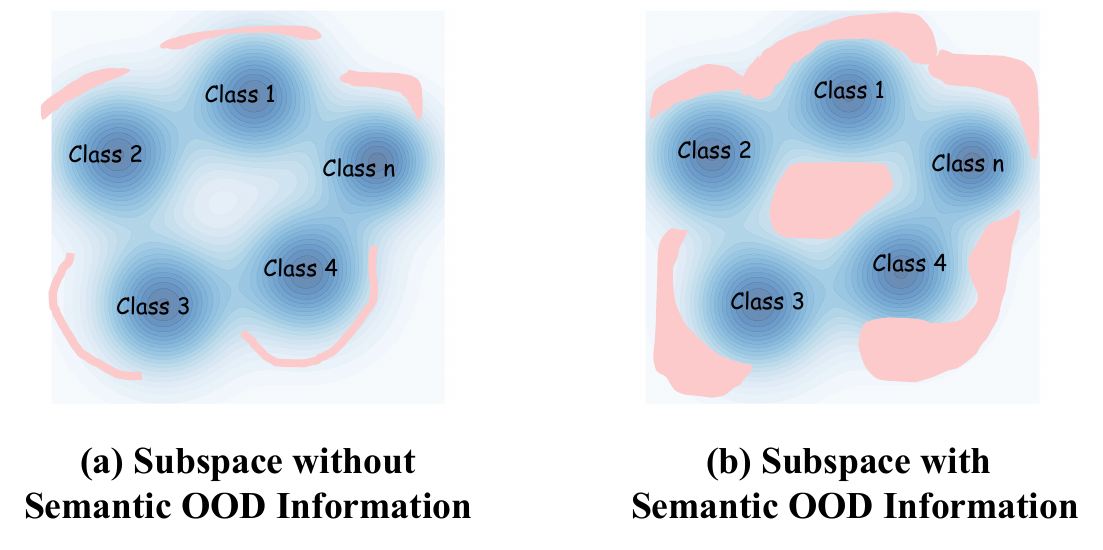}  
    \caption{Comparison of subspaces of methods without semantic OOD information and our proposed CFC, which incorporates such information. Blue regions denote ID subspaces; other regions show OOD subspaces. CFC provides larger embedding space (pink), enabling direct OOD identification.}
    \label{fig-example}
\vspace{-0.1in}
\end{figure}
While these approaches have mitigated the problem of OOD detection to some extent, they encounter several challenges, as follow as: i) To enable OOD detection, they require the use of a large number of synthetic/auxiliary OOD samples during training, imposing significant computation cost. ii) {Using generated samples or auxiliary training data may not accurately reflect real-world OOD variations. As such, these approaches may lack true semantic understanding and risk overfitting to specific datasets pairs.} iii) {Without semantic OOD samples that are realistic and meaningful,} they fail to accurately represent the true OOD space, resulting in a small subspace and sharp boundaries for OOD detection, as illustrated in Figure \ref{fig-example} (a). iv) More critically, these methods often group multiple unknown classes into a single OOD category, 
which significantly reduces their utility. In real-world scenarios, distinguishing between various unknown classes is crucial for informed decision-making, efficient data utilization, and performance in high-stakes applications such as medical diagnosis, autonomous driving, and fraud detection. For instance, in financial networks, grouping diverse fraudulent behaviors—such as phishing attacks, insider threats, and money-laundering schemes—into a single unknown category, oversimplifies their distinctions, hindering the nuanced understanding required for effective mitigation.
This challenge raises a fundamental question: \emph{Can we develop a comprehensive classification approach that seamlessly classifies both known and unknown classes, without requiring labeled samples for OODs?}


Accurate OOD classification poses significant challenges due to the uncertainty surrounding potential OOD labels. Specifically, the OOD domains—whether proximal or distant from the ID domains—are unknown, and even more so the number of unknown classes. 
To overcome this, we map the graph into text space and propose a Coarse-to-Fine Classification (CFC) approach to explore the OOD label space. 
In the first step, leveraging the expert knowledge and reasoning capabilities of LLMs, we design a coarse classifier by creating LLM prompts to detect OOD samples on the test set without prior OOD information and generate potential outlier class labels. 
{The identified OOD data from the coarse classifier provides a potential OOD space with semantic information (i.e., possible meaningful and real-world outlier classes, such as topics of papers, news domains, etc.).} Using these noisy coarse OOD data, we proceed to the second step: constructing a GNN-based fine classifier to further detect OOD samples and perform ID classification. More specifically, we use a label propagation method to remove falsely identified OOD samples, and an improved manifold mixup method \cite{verma2019manifold} for OOD data augmentation, enabling us to effectively predict the distribution of novel classes. Unlike approaches that rely on additional synthetic or auxiliary data for training, CFC captures semantic OOD samples, reducing the distribution discrepancy \cite{wang2023learning} between the training data and real OOD data. As a result, CFC constructs a larger OOD subspace with a smoother boundary for OOD detection, as shown in Figure \ref{fig-example} (b). {An empirical example demonstrating clearer boundaries with more semantic OOD samples is provided in Appendix.} The advantages of semantic OOD are further demonstrated by the improved OOD detection performance achieved with a small number of semantic OOD samples. Finally, we achieve OOD classification using LLM prompts designed with a post-processed OOD label space. It is important to note that we do not have any true OOD label information. Our contributions can be summarized as follows:

\begin{itemize}
    \item Recognizing the critical need for accurately distinguishing between various unknown classes to ensure safety and enable effective decision-making in unpredictable environments, we introduce a novel challenge in the open-world setting: \textbf{OOD classification} on graphs. This task involves not only detecting OOD samples but also classifying them into their respective unknown classes, thereby extending the scope of traditional OOD detection.
    \item We propose a general coarse-to-fine classification method that integrates semantic OOD samples and a potential OOD label space, enabling the model to effectively perform both ID and OOD classification for the open-set graph node classification problem.
    \item Our CFC method demonstrates strong performance, achieving up to a 70\% improvement in graph OOD classification and a 10\% improvement in OOD detection compared to baseline methods. Furthermore, the proposed CFC framework offers a flexible and effective open-set classification solution that can be easily applied to text data.
\end{itemize}

\section{Related Works}


Open-set classification, which identifies unknown classes while classifying known classes, has been widely studied in images and text.
Representative OOD detection methods can be broadly categorized into post-hoc detection \cite{hendrycks2016baseline,liu2020energy,park2023nearest}, generative model-based approaches \cite{cai2023out,kirichenko2020normalizing,nalisnick2018deep,neal2018open}, and outlier exposure techniques \cite{hendrycks2018deep,wang2023learning,hu2021uncertainty}. Post-hoc models employ various scoring functions \cite{liang2017enhancing, sun2022dice,zhu2023unleashing,liu2020energy,huang2021importance} to identify OOD samples, but struggle when test label spaces differ from training, often requiring costly retraining. Generative methods and outlier exposure approaches attempt to incorporate synthetic OOD samples or auxiliary data to train models for OOD detection. However, these generated OOD samples or auxiliary data often fail to accurately represent true OOD samples, limiting their effectiveness.

Recently, there has been growing interest in graph open-set classification \cite{wu2020openwgl,um2025spreading,chen2025decoupled,ma2024revisiting}.
Many classification-based methods \citep{song2022learning,yang2023graph,wu2023energy,ma2024revisiting} have been proposed to utilize structural information for ID classification and OOD detection. However, these methods are often time-consuming and lack flexibility. {To improve efficiency, several works \cite{gong2024energy,yang2025bounded,wu2023energy,wang2025gold} employ energy-based propagation schemes to detect OOD samples.} Additionally, \citet{zhang2023g2pxy} attempted to generate synthetic samples to approximate the OOD space. However, the generated sample distribution often fails to accurately reflect the true OOD distribution, causing these methods to struggle with identifying challenging OOD samples. Importantly, existing open-set classification methods, including node-level \citep{bao2024graph,gong2024energy,zhang2024conc,zhang2024rog_pl} and graph-level \citep{yin2024dream,shen2024optimizing,guo2023data,liu2023good}, typically identify multiple unknown classes as one OOD label. In this paper, we propose a more challenging problem: OOD classification, which involves identifying multiple classes. 
More related works on LLMs for text-attributed graphs can be found in Appendix.


\section{Problem Definition and Preliminaries}

We study open-set node classification in graphs. Given a graph $\mathcal{G} = (\mathcal{V}, \mathcal{E})$ with node set $\mathcal{V}$ and edge set $\mathcal{E}$, let $|\mathcal{V}| = N$. Each node $v_i \in \mathcal{V}$ has a feature vector $\boldsymbol{x}_i \in \mathbb{R}^{d}$, forming a feature matrix $\mathbf{X} \in \mathbb{R}^{N \times d}$, and a label vector $y_i \in \{0,1\}^{C}$, where $C$ is the number of ID classes. Node connections are represented by the adjacency matrix $\mathbf{A}$, where $\mathbf{A}_{ij} = 1$ if $(v_i, v_j) \in \mathcal{E}$, otherwise $\mathbf{A}_{ij} = 0$.  
We consider that the full node set $\mathcal{V}$ is partitioned into training set $\mathcal{V}_\text{train}$, validation set $\mathcal{V}_\text{val}$, and test set $\mathcal{V}_\text{test}$. In a typical closed-set node classification task on graph $\mathcal{G}$, with an ID label space $\mathcal{Y}=\{1, \ldots, C\}$,  GNN models predict each node in the test set with a certain ID class in $\mathcal{Y}$. 

\textbf{OOD detection in open-set node classification problem.}  In open-world scenarios, the test set may contain unknown class nodes whose labels fall outside the ID label space. Given a set of ID training samples $T=\left\{\left(x_1, y_1\right), \ldots,\left(x_n, y_n\right)\right\}$, the goal of OOD detection is to learn a $(C+1)$-class classifier $f_{C+1}$ using $T$. This classifier should be capable of: (1) classifying ID samples into their respective $C$ ID classes, and (2) identifying OOD samples as belonging to a single OOD class.

\textbf{OOD classification in open-set node classification problem.} In this paper, we extend the challenge from OOD detection to OOD classification, formulating it as a comprehensive open-set classification problem. Specifically, we aim to learn a $(C+u)$-class classifier $f_{C+u}$ using $T$ to classify (1)  ID samples into the corresponding $C$ ID classes, and (2) OOD samples into $u$ distinct OOD classes. Notably, $u$ is not predefined in the open-set scenario.

\begin{figure*}[]
	\centering
	\includegraphics[width=6in]{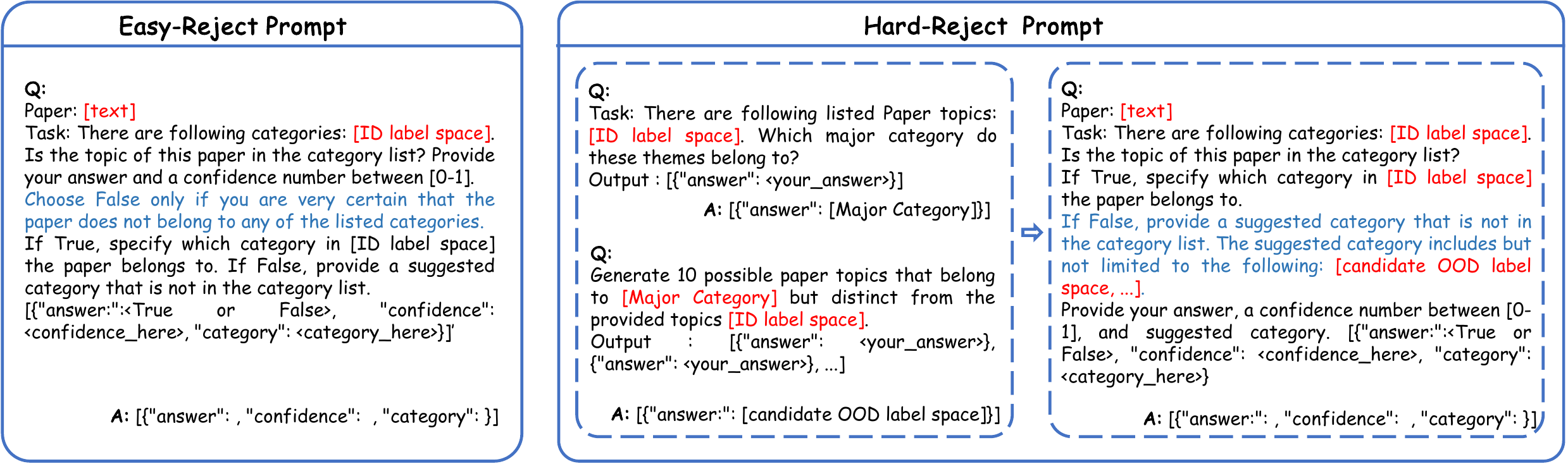}
        \vskip -0.05in
	\caption{ LLM prompts for Easy-Reject and Hard-Reject OOD detection include both Q(uestion) and A(nswer) contents. The inputs are [text] (describing the graph node) and [ID label space] (a list of ID categories, e.g., [machine learning, neural networks, ...]). For Hard-Reject OOD detection, we first determine the [Major Category] of ID classes and the [candidate OOD label space], then use [text], [ID label space], and [candidate OOD label space] for OOD detection and category generation.
	}
	\label{fig-prompt}
    \vskip -0.1in
\end{figure*}

\section{Methods}

To address the challenges of OOD classification, we must tackle two critical questions: (1) How can we approximate the OOD space without labeled information? (2) How can we derive meaningful outlier class labels? 

In this paper, we propose a Coarse-to-Fine open-set Classification (CFC) framework to progressively achieve advanced OOD classification.
First, we design LLM-based prompts specifically tailored for coarse-grained graph node OOD identification. In this step, we leverage the expert knowledge and reasoning capabilities of LLMs to detect OOD samples relevant to the test domain and generate a candidate OOD label space. 
Next, based on the semantic OOD samples identified by the LLM, we introduce a GNN-based fine-grained classifier for ID classification and precise OOD detection. This step enhances granularity by denoising and OOD data augmentation. Furthermore, we provide a theoretical analysis demonstrating the benefits of integrating semantic OOD samples and the refined augmentation method, which expand the OOD subspace and smooth decision boundaries. Finally, we conduct OOD classification by employing LLM prompts with the refined OOD label space.

\subsection{A Coarse-Classifier with LLMs}
\label{llm-coarse}
Large Language Models, with their extensive knowledge, have demonstrated impressive zero-shot and few-shot capabilities, particularly for node classification tasks on text-attributed graphs (TAGs) \cite{chen2023label,guo2023gpt4graph,chen2024exploring}, where each node and edge in the graph is associated with a text sentence. In an open-set setting, we explore the OOD space by leveraging the capabilities of LLMs. Using ID labels from the training set, the LLM is employed to predict whether the label of a test node belongs to the provided ID label space, acting as a binary classifier to differentiate between ID and OOD samples.

According to ID label space, we categorize identification tasks into two types: \textbf{Easy-Reject} and \textbf{Hard-Reject}, as described below. We then elaborate on the corresponding LLM prompts designed to facilitate confidence-aware OOD identification and to generate the potential OOD label space.

\textbf{Easy-Reject.}
This refers to scenarios where the ID classes in the label space contain a small proportion of their respective major categories, making it easier to reject ID samples as OOD class. Building on the existing ID label space, we prompt the LLM to determine whether the label of the input test node belongs to the provided ID classes using confidence-aware prompts \cite{chen2023label}.
The confidence score associated with this identification is essential, as LLM annotations, similar to human annotations, can exhibit a degree of label noise. This confidence score helps assess the quality of detection and filter out noisy labels. Since these samples are likely to be rejected as OOD, we design the LLM prompt with a restriction: annotate samples as OOD only when the LLM is highly confident. If the test node is identified as an ID sample, we prompt the LLM to provide its category within the specified ID label space. Otherwise, the LLM will offer an outlier class label beyond the ID label space. Finally, we obtain the label (ID or OOD), the LLM's confidence score, and the category for each test sample, as illustrated in Figure \ref{fig-prompt}.

\paragraph{Hard-Reject.}
This refers to cases where the ID classes in the label space encompass a large proportion of their respective main categories, making it easier to accept OOD samples as ID ones. Building upon the existing ID label space, we first guide the LLM to summarize these classes and identify their respective major categories. Next, we prompt the LLM to provide possible outlier class labels that fall within the major categories but are not included in the ID class labels, creating a candidate OOD label space. Subsequently, input with the candidate OOD classes, we use the LLM to determine the label of the input test node, generate a confidence score, and provide a predicted category (as illustrated in Figure \ref{fig-prompt}).

In general, Easy-Reject is used for small coverage and far-OOD cases, while Hard-Reject is used for large coverage and near-OOD cases.

\subsection{GNN-based Fine-Classification}
\label{gnn-fine}
Assume we have obtained a coarse-grained OOD set $\mathcal{V}_{\text{ood}}$ through LLM-based OOD detection. {The samples in $\mathcal{V}_{\text{ood}}$ are semantic OODs with potential categories, and are closely aligned with the true test OOD space, providing a more representative and structured foundation for OOD detection.}

We further construct a GNN-based classifier with 
$(C+1)$ labels to perform ID classification and OOD detection. Given that the OOD samples identified by the LLM may contain some noise (i.e., misclassified ID samples) or be insufficient in number, we employ a label propagation method for denoising and utilize an improved mixup method \cite{verma2019manifold,han2022g} for OOD data augmentation.

\paragraph{Denoising.} We correct falsely identified OOD samples from LLM-based OOD detection using a label propagation method. We assume the initial label matrix $\mathbf{Y}^{l(0)}=[y_1^{l(0)}, y_2^{l(0)}, \cdots, y_N^{l(0)}]$ consists of one-hot label indicator vectors for ID training nodes in $\mathcal{V}_{\text{train}}$ and OOD nodes in $\mathcal{V}_{\text{ood}}$, while having zero vectors for unlabeled nodes. By propagating the labels with the normalized adjacency $\mathbf{D}^{-1}\mathbf{A}$, the $k^{th}$ iteration of label propagation \cite{zhu2005semi,wang2020unifying} is formulated as
$\mathbf{Y}^{l(k)}=\mathbf{D}^{-1}\mathbf{A} \mathbf{Y}^{l(k-1)}$.
At each iteration, the ID training samples are reset to their initial labels: $y_i^{l(k)}=y_i^{l(0)}, \forall i \in \mathcal{V}_{\text{train}}$. This is to maintain the label information of the ID training nodes so that the other nodes do not overpower the original labeled ones, as the initial labels would otherwise fade away.

After $K$-order label propagation, we can obtain the label matrix $\mathbf{Y}$. For candidate OOD samples, their labels are updated using the maximum probability in $Y^{K}$. We discard OOD samples that are predicted as ID in $\mathcal{V}_{\text{ood}}$, and achieve new OOD set $\mathcal{V}^{'}_{\text{ood}}$.

\paragraph{OOD Data Augmentation.}
We consider the practical case in which LLM just identifies a small number of samples as OOD. Having a sufficient number of semantic OOD samples is crucial for representing the OOD space and improving open-set classification. How to obtain more stable OOD samples? Manifold mixup \cite{verma2019manifold}, as a data augmentation method, has been theoretically and empirically shown to improve
the generalization and robustness of deep neural networks for images, by training neural networks on linear combinations of hidden representations of training examples. In this work, we extend manifold mixup to augment OOD data for improved performance.

For a well-trained classifier, features of nodes that belong to the same classes are close to each other, while those from different classes are distant. Ideally, clear boundaries separate the different classes. Since nodes whose features are close to the boundaries are more likely to be less representative to their own classes, we generate OOD samples using nodes near the boundary regions. 

We collect $K$ nodes with low classification confidence in the training set. Then, the manifold mixup is applied on these near boundary nodes and the center of the OOD samples in $\mathcal{V}^{'}_{\text{ood}}$ as
\begin{equation}
\left\{\begin{array}{l}
\tilde{x}_i=\alpha \boldsymbol{h}_i^k + (1-\alpha) \boldsymbol{h}_{c}^k, i \leq K \\
\tilde{y}_i=C+1
\end{array}\right.
\label{mixup}
\end{equation}
where $\boldsymbol{h}^{k}=\operatorname{GNN}\left(\mathbf{A}, \boldsymbol{h}^{k-1}\right)$ is the hidden embedding with a GNN encoder, $\boldsymbol{h}_{c}^k$ is the center embedding of OOD samples, $\alpha > 0$ is a hyperparameter to control the distance between the generated samples and the existing OOD samples. Finally, we obtain an augmented OOD set $\mathcal{V}_{\text{ood}}^{a} = \{ \mathcal{V}^{'}_{\text{ood}}, \mathcal{V}_{a} \}$ where $\mathcal{V}_{a}$ is the generated OOD set.

\paragraph{ID Classification and OOD Detection.} We train a GNN-based classifier $f_{C+1}$ on training set $\mathcal{V}_{\text{train}}$ and the augmented OOD set $\mathcal{V}_{\text{ood}}^{a}$ for ID classification and OOD detection. Taking GCN as an example, a two-layer GCN model can be formulated as
\begin{equation}
\boldsymbol{Z}=\operatorname{softmax}\left(\hat{\mathbf{A}} \operatorname{ReLU}(\hat{\mathbf{A}} \mathbf{X} \mathbf{W}^{(0)}) \mathbf{W}^{(1)}\right),
\label{eq:gcn}
\end{equation}
where, $\boldsymbol{Z}$ is the GCN's output predictions, $\hat{\mathbf{A}}=\hat{\mathbf{D}}^{-\frac{1}{2}}\left(\mathbf{A}+\mathbf{I}_{n}\right) \hat{\mathbf{D}}^{-\frac{1}{2}}$ is the normalized $\mathbf{A}+\mathbf{I}_{n}$ matrix by the degree matrix $\hat{\mathbf{D}}$, and  $\mathbf{W} = (\mathbf{W}^{(0)}, \mathbf{W}^{(1)})$ are the weights of the two-layer GCN model. For graph node classification, the objective function $\mathcal{L}$ is
\begin{equation}
\mathcal{L} \;=\; -\frac{1}{|\mathcal{V}_{\text{train}} \cup \mathcal{V}_{\text{ood}}^{a}|}
\sum_{v_i \in \mathcal{V}_{\text{train}} \cup \mathcal{V}_{\text{ood}}^{a}}
\boldsymbol{y}_i^\top \log(\boldsymbol{z}_i),
\end{equation}
where $\boldsymbol{y}_i$ and $\boldsymbol{z}_i $ are the label and prediction of node $v_i$. We use the trained GNN-based open-set classifier to predict test labels and obtain the final predicted OOD set, denoted as $\mathcal{V}_{\text{ood}}^{f}$.  

\subsection{OOD Classification}
\label{ood-class}

\begin{figure}[]  
    \centering
    \includegraphics[width=0.37\textwidth]{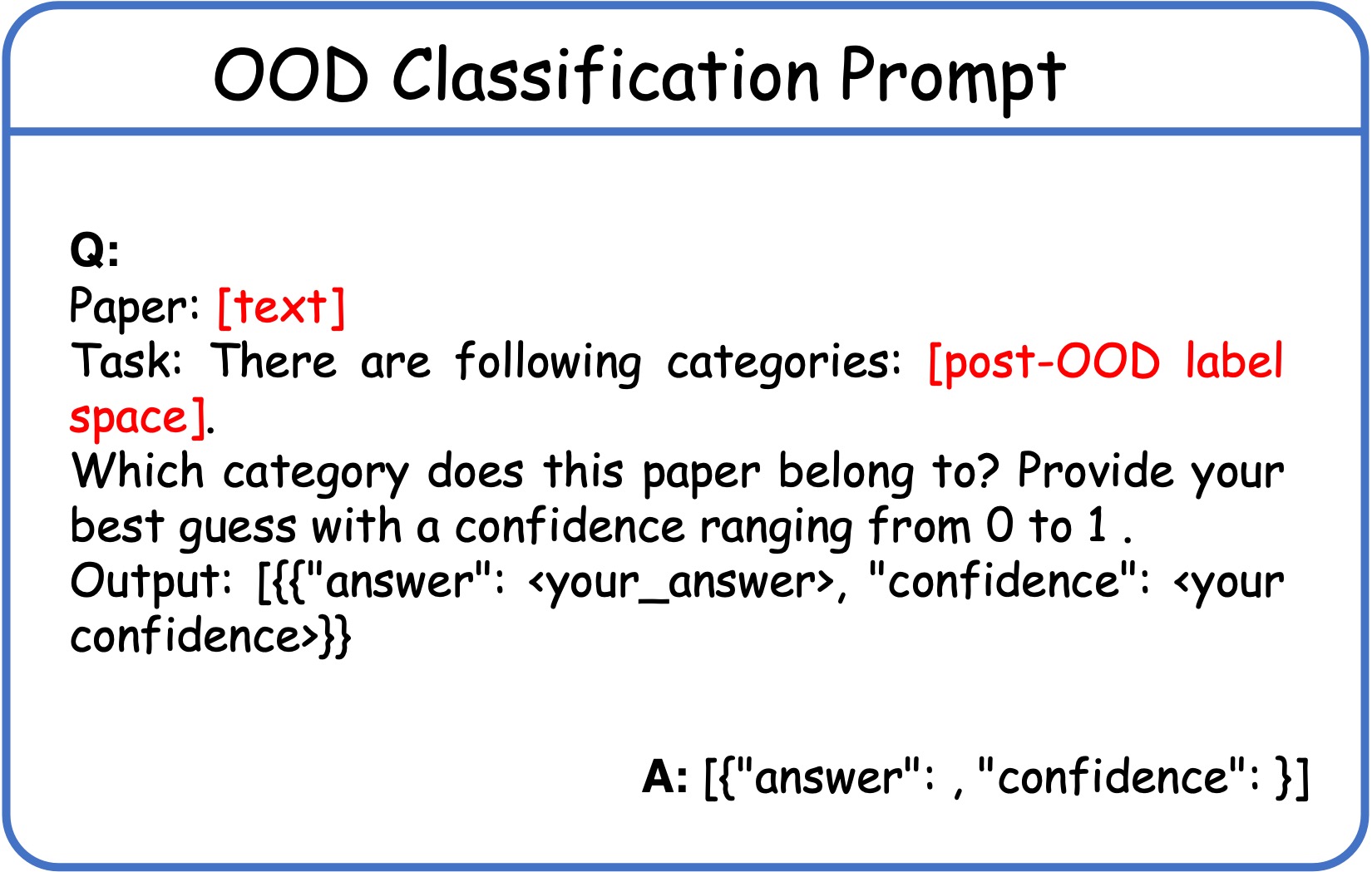}  
    \caption{LLM prompts with [text] and [post-OOD label space] for OOD classification.}
    \label{fig-oodclass}
\end{figure}
After detecting the OOD samples, we proceed with OOD classification for the nodes in $\mathcal{V}_{\text{ood}}^{f}$, by leveraging the potential OOD label space (consisting of the possible OOD categories) discussed in subsection Coarse-Classifier with LLMs.

We employ similarity measures (such as word-level or semantic-level comparisons, e.g., TF-IDF \citep{ramos2003using}) to merge similar categories and filter out categories with too few samples. After post-processing, we obtain a set of OOD categories, $ \{l_1, l_2, \cdots, l_u\}$, forming the post-OOD label space. We then use LLMs to generate annotations for the OOD samples in $\mathcal{V}_{\text{ood}}^{f}$ based on this post-OOD label space, as illustrated in Figure \ref{fig-oodclass}.


\section{Theoretical Analysis}
In CFC, we integrate augmented semantic OOD samples by proposing a method to mix the hidden embeddings of samples from ID classes and coarse OOD samples obtained through coarse-grained OOD detection. We theoretically demonstrate the advantages of this approach in extending and flattening the OOD subspace (as shown in Figure \ref{fig-example} (b)), which leads to improved OOD detection.

\textbf{Definition 3.1} (Space Dimension)\textbf{.}
\textit{{Given a hypothesis space $\mathcal{H}$, the space dimension is the rank of the matrix formed by the set of vectors that span the subspace.}}

\textbf{Assumption 3.1} (Feature Space Dimension)\textbf{.}
\textit{{Given a feature space $\mathcal{X} \subset \mathbb{R}^d$, assume that the features of $\mathcal{X}$ belong to $C$ distinct classes. Then, the dimension of the feature space is $d - C$.}}

\textbf{Theorem 3.1}
\textit{{Given an open-set classification task with an ID label space $\mathcal{Y}_{\text{id}} = \{ y_1, y_2, \dots, y_C \}$ and an OOD label $y_{\text{ood}}$, assume the existence of the ID feature space $\mathcal{H}$ and the OOD space ${\mathcal{H}}^{'}$ related to practical test domain. The proposed CFC, trained on both ID samples and semantic OOD samples related to the OOD space ${\mathcal{H}}^{'}$, forms a larger subspace with dimension $dim(\mathcal{H} + \mathcal{H}^{'}) - (C + 1)$, compared to general methods that lie within $dim(\mathcal{H}) - (C + 1)$. Consequently, CFC results in a smoother and flatter decision boundary for OOD detection.}}

{A theoretical analysis of how embedding mixup further expands the representation subspace and influences the classifier’s decision boundary, and the proof of Theorem 3.1, is provided in Appendix.}

\section{Experiments}

In this section, we evaluate the performance of our proposed CFC method for graph node classification in an open-set setting by investigating the following questions:
\textbf{Q1.} How does the performance of CFC compare to other classification methods?
\textbf{Q2.} What is the effect of different prompts and LLMs on coarse-grained OOD detection?
\textbf{Q3.} How do denoising and OOD data augmentation in fine-grained OOD detection affect the performance of CFC?
\textbf{Q4.} What is the impact of the potential OOD label space on OOD classification?
Additionally, we demonstrate that the proposed CFC framework is versatile and can be extended to the text domain in Appendix.


\paragraph{Experimental Settings.}
{In this paper, we utilize widely used graph datasets from different domains—textual graphs (Cora \cite{mccallum2000automating}, Citeseer \cite{giles1998citeseer}, WikiCS \cite{mernyei2020wiki}, and DBLP \cite{ji2010graph}) and non-textual graphs (Amazon-Computer and Amazon-Photo \cite{ni2019justifying})—for open-set node classification.} The statistics for these datasets are presented in Appendix. For each dataset, multiple classes are designated as out-of-distribution (OOD) classes (i.e., $u>=2$), while the remaining classes are considered in-distribution (ID) classes.
Further details for the case with  $u = 2$ and $u = 3$ can be found in the Appendix. For the ID classes, 50\% of the nodes are sampled for training. The remaining ID samples and all OOD samples are split as 40\% for validation and 60\% for testing.

For all datasets, we adopt the text-attributed graph versions from \cite{chen2023label, liu2023one, chen2024text}. We use four popular large language models (LLMs), including e5-large-v2 (e5, \cite{wang2022text}), Sentence Transformer (ST, \cite{reimers2019sentence}), Llama2-7b and Llama2-13b \cite{touvron2023Llama}, to generate embeddings as original input features. We utilize GPT-4o \citep{achiam2023gpt} for detecting OOD samples and generating potential outlier class labels for coarse-grained OOD identification, as well as for the final OOD classification. 
In the coarse classifier with LLM, we apply a confidence threshold of 0.7 to identify OOD samples, \textcolor{black}{using Easy-Reject for Cora, DBLP, and WikiCS, and Hard-Reject for Citeseer, Computer, and Photo.}
In fine-grained classification, we generate 100 OOD samples for text datasets, and over 2000 for Computers and Photo datasets using improved manifold mixup. The hyperparameter analysis of $\alpha$ used to balance ID features and OOD features in Eq.\eqref{mixup} can be found in Appendix. We compare our method with popular closed-set classification methods and state-of-the-art open-set classification methods, including {GCN\_Poser} \cite{zhou2021learning}, ${G}^2Pxy$ \cite{zhang2023g2pxy}, GNNSafe \cite{wu2023energy}, NodeSafe \cite{yang2025bounded}, and GOLD \cite{wang2025gold}.








\begin{table*}[]
    \centering
    \caption{Comparison of different methods for ID classification and OOD detection across four datasets with two OOD classes. Note that CFC (w/o D/M) refers to the CFC framework without the Denoising and Manifold Mixup data augmentation techniques. CFC uses GCN as the backbone for fine classification. (mean accuracy (\%) and standard deviation over 5 different runs).}
    \begin{center}
        \vskip -0.1in
    \renewcommand\arraystretch{1.1}
\resizebox{\textwidth}{!}{%
    \begin{tabular}{l|ccc|ccc|ccc|ccc}
        \hline
        \multirow{2}{*}{Methods} & \multicolumn{3}{c|}{Cora} & \multicolumn{3}{c|}{Citeseer} & \multicolumn{3}{c|}{WikiCS} & \multicolumn{3}{c}{DBLP}\\
        \cline{2-13}
                                 & ID &  OOD & overall  & ID & OOD & overall & ID & OOD & overall & ID & OOD & overall\\
        \hline
        GCN\_softmax         & $90.25_{\pm0.85}$  & 0.0 & $62.76_{\pm0.72}$ & $76.15_{\pm2.25}$ & 0.00   & $38.60_{\pm1.18}$ & $73.67_{\pm0.56}$ & 0.0 & $53.01_{\pm0.32}$ & $91.11_{\pm1.34}$ & 0.0  & $56.62_{\pm0.82}$ \\
        GCN\_sigmoid         & $90.64_{\pm0.54}$ & 0.0 & $63.03_{\pm0.53}$ & $76.07_{\pm1.17}$ & 0.00 & $38.56_{\pm0.58}$ & $60.18_{\pm1.17}$ & 0.00 & $43.30_{\pm0.68}$ & $91.54_{\pm0.58}$ & 0.00 & $56.89_{\pm0.50}$   \\        
        GCN\_softmax\_$\tau$   & $81.13_{\pm2.81}$    & $66.98_{\pm9.87}$ & $76.84_{\pm1.36}$  & $57.64_{\pm4.27}$ & 81.97 & $69.60_{\pm1.21}$ & $41.91_{\pm7.75}$  & $58.84_{\pm12.31}$  & $46.68_{\pm2.75}$ & $79.15_{\pm4.52}$ & $45.14_{\pm6.59}$ & $66.28_{\pm 0.73}$ \\
        
        GCN\_sigmoid\_$\tau$    & $85.17_{\pm2.18}$  & $62.18_{\pm5.30}$ & $78.16_{\pm1.02}$ & $67.52_{\pm1.46}$ & $75.41_{\pm2.31}$ & $71.41_{\pm1.01}$ & $54.13_{\pm0.64}$ & $67.03_{\pm1.66}$ & $57.74_{\pm0.67}$  & $71.11_{\pm7.32}$ & $61.54_{\pm7.92}$ & $66.31_{\pm1.25}$  \\
        GCN\_PROSER              & $84.21_{\pm0.73}$ & $71.18_{\pm1.75}$ & $80.65_{\pm 0.73}$ & $71.25_{\pm5.15}$ & $76.01_{\pm3.65}$ & $72.74_{\pm1.57}$ & $48.13_{\pm0.99}$ & $44.19_{\pm1.63}$  & $46.74_{\pm0.64}$ & $63.01_{\pm10.88}$ & $68.75_{\pm14.98}$ & $65.18_{\pm1.27}$ \\
        ${G}^2Pxy$   & $85.65_{\pm1.06}$ & $72.46_{\pm1.58}$ & $81.63_{\pm0.77}$ & $71.52_{\pm1.43}$ & $77.30_{\pm2.63}$ & $74.36_{\pm1.30}$ & $58.40_{\pm0.84}$ & $57.09_{\pm1.50}$ & $58.03_{\pm0.45}$ & $65.88_{\pm1.03}$ & $62.63_{\pm0.98}$ & $64.65_{\pm1.23}$ \\
        GNNSafe & $79.06_{\pm1.53}$ & $62.92_{\pm9.28}$ & $74.14_{\pm3.11}$ & $71.43_{\pm1.37}$ & $36.18_{\pm5.08}$ & $53.64_{\pm2.76}$ & $83.56_{\pm2.85}$ & $73.29_{\pm13.91}$ & $79.59_{\pm4.21}$ & $93.85_{\pm0.47}$ & $47.25_{\pm2.34}$ & $76.21_{\pm0.55}$ \\
        NodeSafe & $87.93_{\pm1.13}$ & $80.63_{\pm4.76}$ & $85.71_{\pm0.82}$ & $73.97_{\pm 1.88}$ & $53.81_{\pm 4.43}$ & $64.03_{\pm 1.99}$ & $86.66_{\pm0.42}$ & $42.99_{\pm 3.31}$ & $70.03_{\pm 1.51}$ & $94.52_{\pm 0.28}$ & $42.50_{\pm3.33}$ & $74.83_{\pm 1.10}$\\
        GOLD & $87.48_{\pm1.07}$ & $66.54_{\pm 2.84}$ & $81.11_{\pm 1.35}$ & $70.33_{\pm1.44} $ & $37.75_{\pm6.39}$ & $53.26_{\pm3.02}$ & $73.60_{\pm1.93}$  & $32.83_{\pm14.98}$ & $58.12_{\pm6.68}$ & $94.05_{\pm0.41}$ & $43.68_{\pm 1.44}$ & $74.98_{\pm 0.61}$\\
        GTP-4o & 72.23 & 58.08 & 68.62 & 46.82 & 48.18 & 47.50 & 67.33 & 67.65 & 67.43 & 84.26 & 56.78 & 66.11  \\
        \hline
        CFC (wo / D/M) & $85.44_{\pm0.20}$ & $94.50_{\pm0.31}$ & $88.20_{\pm0.20}$ & $76.59_{\pm2.12}$ & $72.92_{\pm2.63}$ & $74.77_{\pm0.53}$ & $79.18_{\pm0.13}$ & $81.76_{\pm0.16}$ & $79.91_{\pm0.10}$ & $75.18_{\pm2.84}$ & $87.62_{\pm1.11}$ & $83.40_{\pm0.27}$ \\
        CFC (wo / M) & $88.63_{\pm0.53}$ & $91.75_{\pm1.57}$ & $89.58_{\pm0.35}$ & $79.29_{\pm2.10}$ & $67.49_{\pm1.50}$ & $73.44_{\pm0.77}$ & $80.19_{\pm1.04}$ & $77.13_{\pm0.68}$ &  $79.32_{\pm0.65}$ & $76.25_{\pm3.76}$ & $85.00_{\pm1.61}$ & $82.03_{\pm0.38}$ \\
        CFC (wo / D) & $86.23_{\pm0.64}$ & $94.91_{\pm0.85}$ & $88.87_{\pm2.15}$ & $72.40_{\pm1.52}$ & $\textbf{83.00}_{\pm1.66}$ & $\textbf{77.65}_{\pm0.86}$ & $75.15_{\pm0.24}$ & $\textbf{89.95}_{\pm0.21}$ & $79.34_{\pm0.19}$ & $75.37_{\pm1.79}$ & $\textbf{88.37}_{\pm0.29}$ & $83.96_{\pm0.47}$\\
        CFC & $87.49_{\pm0.80}$ &  $\textbf{95.74}_{\pm0.67}$ & $\textbf{90.00}_{\pm0.37}$ & $73.92_{\pm0.97}$ & $80.57_{\pm1.69}$ & $77.21_{\pm0.68}$ &  $80.19_{\pm0.73}$ & $81.89_{\pm0.53}$  & $\textbf{80.44}_{\pm0.10}$ & $78.47_{\pm1.01}$ & $86.89_{\pm0.71}$  & $\textbf{84.03}_{\pm0.15}$ \\
        \hline
    \end{tabular}
}
    \end{center}
    \vskip -0.1in
    \label{main-ood2}
\end{table*}

\subsection{Comparison With Other Node Classification Methods}

We conducted two tasks: the traditional open-set graph node classification (ID classification and OOD detection) and the newly proposed OOD classification (mean accuracy (\%) and standard deviation over 5 different runs).


\subsubsection{OOD Detection}
Table \ref{main-ood2} presents a comparison of closed-set and recent state-of-the-art open-set graph node classification methods for ID classification and OOD detection across four text-attributed graph datasets. Here, CFC adopts e5-large-v2 as the feature encoder. We observe that the proposed CFC method outperforms all other baselines across all datasets by significantly large margins. Specifically, CFC achieves over or around a 10\% improvement over the second-best in terms of overall accuracy on Cora, WikiCS, and DBLP. Furthermore, even without denoising and OOD data augmentation, CFC (wo / D/M) delivers comparable or better results on all datasets compared to other baselines. This suggests that incorporating semantic OOD information related to true OOD domain during training benefits the model, which is a reasonable outcome. Additionally, we observe that the LLM (GPT-4o) can recognize only about half of the OOD samples when provided with the ID label space in most cases, making it unsuitable for high-stakes applications. We report the experimental results of various graph node classification methods under the scenario of three OOD classes in each dataset in Appendix. Since DBLP has only four classes, we only report results on the other three datasets. Our proposed CFC outperforms the other baselines by a significant margin (over 10\% overall accuracy) across all datasets. {The effectiveness of CFC on non-textual graph datasets is demonstrated in Table \ref{main-ood2-nontextual}.}



\begin{table}[]
    \centering
    \caption{{Comparison of different methods for ID classification and OOD detection across two nontextual datasets with two OOD classes.}}
    \renewcommand\arraystretch{1.1}
    \resizebox{0.45\textwidth}{!}{%
    \begin{tabular}{l|ccc|ccc}
        \hline
        \multirow{2}{*}{Methods} & \multicolumn{3}{c|}{Amazon-Computer} & \multicolumn{3}{c}{Amazon-Photo} \\
        \cline{2-7}
                                 & ID &  OOD & overall  & ID & OOD & overall \\
        \hline
        GCN\_softmax         &  $81.87_{\pm 1.77}$ & 0.0 & $41.98_{\pm0.90}$ & $82.46_{\pm 0.45}$ & 0.0 & $70.03_{\pm0.38}$ \\
        GCN\_sigmoid         & $73.85_{\pm 0.55}$ & 0.0 & $37.51_{\pm0.28}$ & $67.55_{\pm0.64}$ & 0.0 & $57.37_{\pm0.54}$ \\        
        GCN\_softmax\_$\tau$   & $81.84_{\pm 1.78}$ & $0.08_{\pm 0.08}$ &$41.60_{\pm 0.90}$ & $82.43_{\pm0.44}$ & $0.60_{\pm0.57}$ & $70.10_{\pm0.36}$\\
        GCN\_sigmoid\_$\tau$    &  $14.15_{\pm12.77}$ & $93.11_{\pm6.60}$ & $52.33_{\pm3.39}$ & $41.20_{\pm1.39}$ & $\textbf{86.66}_{\pm3.22}$ & $47.92_{\pm1.60}$\\
        GNNSafe & $70.62_{\pm1.21}$ & $42.66_{\pm 4.21}$ & $56.86_{\pm 1.01}$ & $69.10_{\pm0.98}$ & $12.85_{\pm5.68}$ & $60.62_{\pm2.54}$ \\
        NodeSafe & $86.73_{\pm 0.52}$ & $66.64_{\pm2.47}$ & $76.90_{\pm1.29}$ & $84.00_{\pm0.58}$ & $48.09_{\pm3.39}$ & $78.58_{\pm0.29}$ \\
        GOLD & $73.18_{\pm3.25}$ & $32.74_{\pm10.86}$ & $52.87_{\pm6.15}$ & $69.58_{\pm 2.64}$ & $3.50_{\pm1.32}$ & $59.61_{\pm2.32}$ \\
        \hline
        CFC & $78.15_{\pm0.96}$ & $\textbf{86.54}_{\pm 1.03}$ & $\textbf{82.28}_{\pm 0.41}$ & $82.81
        _{\pm 0.77}$ & $76.14_{\pm 4.76}$ & $\textbf{81.81}_{\pm 0.07}$\\
        \hline
    \end{tabular}
    }
    \vspace{-0.1in}  
    \label{main-ood2-nontextual}
\end{table}
Table \ref{tab-robust} in Appendix \ref{robust} demonstrates the robustness of CFC under limited ID data.
Table \ref{main-encoder} in Appendix \ref{app-encoders} compares different LLM feature encoders. Overall, e5-large-v2 performs better and more stably in joint training. Additionally, LLMs specialize in different topics—for instance, Llama2 excels in ID topics for Citeseer, while e5 and ST perform well in OOD.

Table \ref{main-ood2} provides detailed classification accuracy for both known (ID) and unknown (OOD) classes. While ID classification performance slightly decreases compared to closed-set methods (e.g., from 90.64\% to 87.49\% on Cora when comparing CFC to GCN\_sigmoid), OOD detection accuracy significantly improves from 0\% to 95.74\%, which is remarkable. Compared to open-set classification methods such as $G^2Pxy$, CFC improves OOD detection accuracy from 72.46\% to 95.74\%, while also enhancing ID classification from 82.65\% to 87.49\% on Cora. The same trend is observed in other datasets. This shows that CFC better separates known and unknown classes by using semantic OOD samples.

\subsubsection{OOD Classification}
Without extra label information, it is evident that existing closed-set and open-set classification methods are not equipped to handle OOD classification when multiple OOD classes are present. Although some methods, like GCN\_Poser and ${G}^2Pxy$, generate OOD samples during training, they struggle to differentiate between various OOD classes without access to OOD label information. Leveraging the annotation capabilities of LLMs and special designed prompt, our proposed CFC method successfully classifies unknown samples into different OOD labels. Notably, using the post-OOD label space and GPT-4o, we achieve accuracies of 69.76\%, 70.30\%, 57.96\%, and 48.45\% on Cora, Citeseer, WikiCS, and DBLP, respectively, in the case of two OOD classes. More OOD classification results with other encoders can be found in Appendix.


\begin{figure*}[!htbp]
	\centering
	\includegraphics[width=6in]{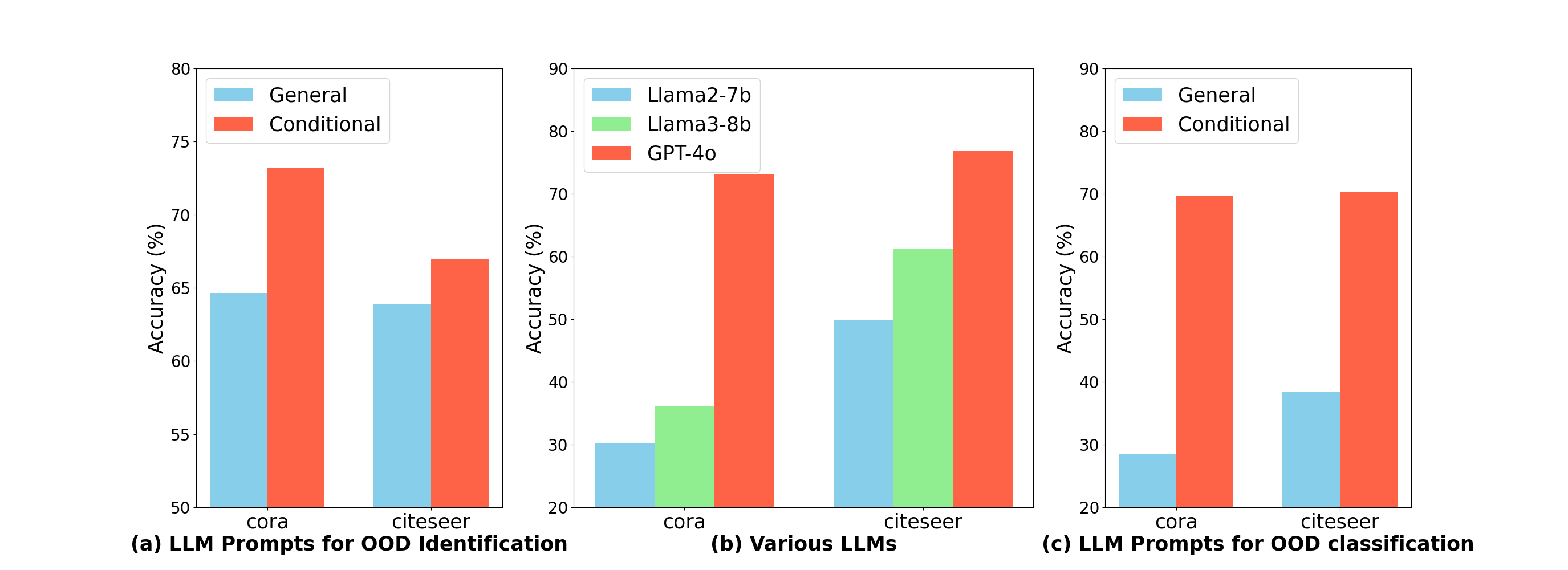}
	\caption{ Ablation study on (a) LLM prompts for OOD identification, (b) Various LLM for OOD identification, and (c) LLM prompts for OOD classification on Cora and Citeseer.
	}
	\label{fig-cora-citeseer}
\end{figure*}

\begin{figure}[]  
  \centering
  \includegraphics[width=0.5\textwidth]{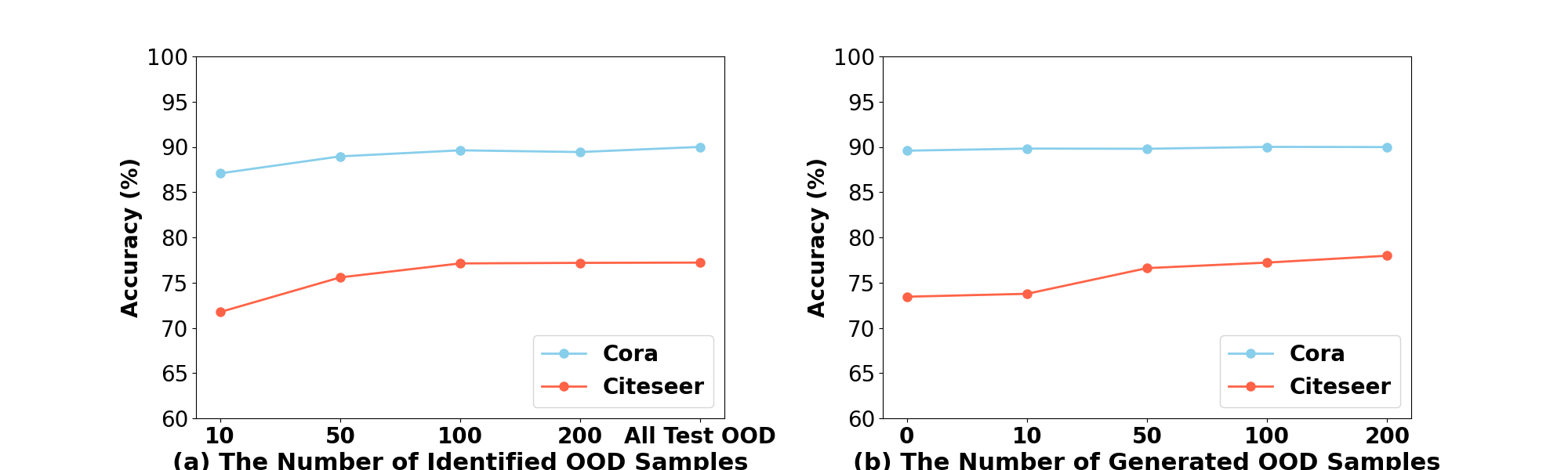}
  \vspace{-0.1in}
  \caption{Study on the effect of (a) the number of identified OOD samples from coarse-grained classification, and (b) the number of generated OOD samples by manifold mixup method for the CFC performance on Cora and Citeseer.}
  \label{fig-num}
  \vspace{-0.2in}
\end{figure}

\subsection{Impact of Different Prompts and LLMs in Coarse-Classifier}


\paragraph{LLM Prompts.} We investigated the effectiveness of using a constraint (rejecting with high confidence for \textbf{Easy-Reject} detection and integrating a candidate OOD label space for \textbf{Hard-Reject} detection) when designing LLM prompts. The rest of the LLM prompt content remains unchanged. Specifically, the \textbf{reject with high confidence} LLM prompt instructs the model to classify an input sample as OOD only when it is highly confident. The LLM prompt with the \textbf{candidate OOD label space} provides the model with additional label choices and encourages consideration of extra labels when making decisions. As shown in Figure \ref{fig-cora-citeseer} (a), without the constraint, OOD performance degrades on AUROC metrics, underscoring the importance of the proposed constraint (Cora utilizes the \textbf{reject with high confidence} LLM prompt, and Citeseer uses the \textbf{candidate OOD label space} LLM  prompt).

\paragraph{Various LLMs.} We conducted experiments with various LLMs to gain a more comprehensive understanding of their ability to detect OOD samples. Specifically, we use Llama (Llama2-7b and Llama3-8b) \cite{touvron2023Llama} and GPT-4o for OOD detection.
Llama models use few-shot learning with one ID and one OOD example; GPT-4o uses zero-shot learning. All prompts follow the constriction strategy. Results on Cora and Citeseer (Figure \ref{fig-cora-citeseer} (b)) show GPT-4o outperforms Llama, with Llama3-8b slightly better than Llama2-7b.

\subsection{Impact of Different Strategies in Fine-Grained Detection}

We conducted ablation studies to evaluate different strategies, including denoising and data augmentation methods for GNN-based fine-grained classification under the case $u=2$. As shown in Tables \ref{main-ood2}, both CFC (wo / D/M) and CFC (wo / D) models, which exclude the denoising, consistently achieve lower ID accuracies across all datasets compared to their counterparts, CFC (wo / M) and CFC with denoising. The CFC (wo / D) model, which employs the improved manifold mixup method to generate more OOD data, demonstrates significant improvement over CFC (wo / D/M) across all datasets. Manifold mixup serves as a regularizer, encouraging the neural network to make less confident predictions on interpolated hidden representations. By integrating ID classes with coarse OOD classes, CFC produces a GNN-based classifier with smoother decision boundaries across multiple representation levels. 
We also found that data augmentation has a greater impact than denoising, since the LLM-designed prompt already helps reduce noise.


In fine-grained detection, the OOD samples identified from coarse-grained detection play a crucial role. Given LLMs’ high cost on large test sets, we evaluate how the number of identified OOD samples affects CFC’s performance. As shown in Figure \ref{fig-num} (a), CFC performs well even with few identified OOD samples, highlighting the advantages of incorporating semantic OOD samples. This demonstrates the scalability of CFC, making it applicable to large-scale test datasets, as evidenced by WikiCS. Additionally, increasing the number of identified OOD samples consistently boosts CFC accuracy on both Cora and Citeseer.

The manifold mixup-based data augmentation method is a key component of CFC. To assess its impact, we examine how the generated OOD samples influence CFC's performance. In Figure \ref{fig-num} (b), we observe the performance varies across different datasets, specifically, an improved performance on Citeseer and stable performance on Cora with an increasing number of generated OOD samples.

\subsection{Impact of OOD Label Space for OOD Classification}

We examine the impact of the post-OOD label space in LLM prompts for OOD classification. Specifically, we evaluate the performance of LLM annotations with and without the potential OOD label space on the predicted OOD samples from fine-grained OOD detection. 
In coarse-grained detection, the LLM generates potential OOD labels using a general prompt without the post-OOD label space as a baseline. For datasets like Citeseer, even when candidate OOD labels are provided, the label space remains broad and comparable to the general prompt.
As shown in Figure \ref{fig-cora-citeseer} (c), using a conditional prompt with a post-processed OOD label space greatly improves prediction accuracy.

\section{Conclusion}

We tackle the open-world challenge of OOD classification with CFC, a coarse-to-fine framework leveraging LLMs. LLM prompts guide coarse OOD detection and build a candidate OOD label space, from which semantic OOD samples are generated. These samples are used to train a GNN-based fine-grained classifier with denoising and data augmentation. A refined LLM prompt then performs OOD classification. Experiments show CFC achieves state-of-the-art OOD detection and strong OOD classification across graph and text domains. We leave graph-level OOD classification for future work and aim to inspire further research in open-set learning.
CFC assumes graph nodes can be described by text; if not, LLMs may struggle with OOD detection and classification. It also depends on the LLM’s knowledge of ID categories. 
{To reduce reliance on large LLMs, fine-tuning smaller models on domain-specific data and integrating retrieval-augmented generation (RAG) into CFC is a promising direction.}

\section*{Acknowledgments}
\textcolor{black}{This work is in part supported by the National Natural Science Foundation of China (Grant No. 62276067) and Australian Research Council Discovery project DP230101534.}

\bibliographystyle{aaai2026}

\clearpage
\appendix

\section{Empirical case study}
\label{app-case}
In Figure \ref{fig-example}, we observe that GNNs can form clear boundaries between ID and OOD samples when more actual OOD data is incorporated, leading to improved OOD detection performance.

\begin{figure}[H]  
    \centering
    \includegraphics[width=3.5in]{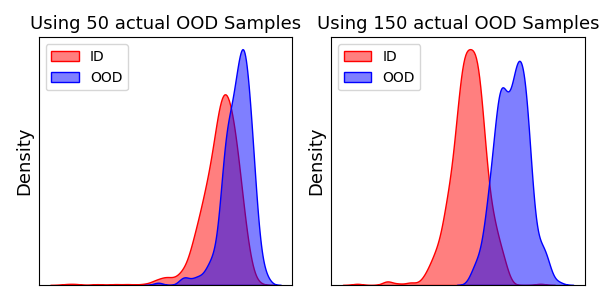}  
    \caption{Comparison of OOD detection on Cora with different number of true OOD samples. Adding actual OOD labels can largely increase the OOD detection performance.}
    \label{fig-example-app}
\end{figure}

\section{More Related Works}
\label{more_works}
\paragraph{LLM for Text-Attributed Graphs.}
Recent progress in applying LLMs to graphs \citep{he2023explanations,guo2023gpt4graph,liu2023one,chen2024text} aims to harness the power of LLMs to enhance the performance of graph-related tasks. LLMs can either be used as predictors \citep{chen2024exploring,wang2024can,ye2023natural}, which directly generate solutions, or as enhancers \citep{he2023explanations}, which leverage the capabilities of LLMs to improve the performance of smaller models with greater efficiency. Additionally, LLMs are explored for graph active learning, serving as annotators \citep{chen2023label} to train efficient models that achieve promising performance without the need for ground truth labels. In this paper, we utilize LLMs for graph node OOD detection and help to address the problem of graph node OOD classification.

\section{Prompts}

In this section, we present the prompts designed for coarse-grained OOD identification and OOD classification. Specifically, Table \ref{cora-prompt} provides an example using the Easy-Reject prompt on Cora, while Table \ref{citeseer-prompt} illustrates the Hard-Reject prompt on Citeseer. Tables \ref{prompt-1} and \ref{prompt-2} offer examples for generating the major category and a candidate OOD label space. Finally, Table \ref{ood-prompt} presents the prompt incorporating the post-OOD label space for OOD classification.

\begin{table}[]
\centering
\caption{Full prompt example with Easy-Reject scenarios for zero-shot coarse classifier on the Cora dataset.}
\label{table:prompt-example} 
    \resizebox{0.45\textwidth}{!}{%
\begin{tabular}{|@{}p{0.9\linewidth}@{}|}
\hline
\textbf{Input:} \\
Paper: \\
A self-adjusting dynamic logic module: This paper presents an ASOCS (Adaptive Self-Organizing Concurrent System) model for massively parallel processing of incrementally defined rule systems in such areas as adaptive logic, robotics, logical inference, and dynamic control. An ASOCS is an adaptive network composed of many simple computing elements operating asynchronously and in parallel. This paper focuses on Adaptive Algorithm 2 (AA2) and details its architecture and learning algorithm. AA2 has significant memory and knowledge maintenance advantages over previous ASOCS models. An ASOCS can operate in either a data processing mode or a learning mode. During learning mode, the ASOCS is given a new rule expressed as a boolean conjunction. The AA2 learning algorithm incorporates the new rule in a distributed fashion in a short, bounded time. During data processing mode, the ASOCS acts as a parallel hardware circuit. \\
\textbf{Task:} \\
There are following categories: \\
{neural\_networks, genetic\_algorithms, theory, reinforcement\_learning, probabilistic\_methods.}\\
Is the topic of this paper in the category list? Provide your answer and a confidence number between [0-1]. \\
\textbf{Choose False only if you are very certain that the paper does not belong to any of the listed categories.} \\
If True, specify which category in \\
{neural\_networks, genetic\_algorithms, theory, reinforcement\_learning, probabilistic\_methods} the paper belongs to. If False, provide a suggested category that is not in the category list. \\
$[ \{ \text{"answer"} : \langle \text{True or False} \rangle, \, 
   \text{"confidence"} : \langle \text{confidence\_here} \rangle, \, 
   \text{"category"} : \langle \text{category\_here} \rangle \} ]$ \\

\textbf{Output:} \\
\hline
\end{tabular}
    }
\label{cora-prompt}
\end{table}

\begin{table}[H]
\centering
\caption{Prompt example for generating major category based on the provided ID label space.}
\label{table:prompt-example}
\begin{tabular}{|@{}p{0.9\linewidth}@{}|}
\hline
\textbf{Input:} \\
\textbf{Task:} \\
There are following listed Paper topics: \\
\{agents, information retrieval, database, \\ artificial intelligence\}.\\
Which major category do these themes belong to? \\
Output : $[\{\text{"answer"}: \langle \text{your\_answer} \rangle \}]$.\\
\textbf{Output:} \\
\hline
\end{tabular}
\label{prompt-1}
\end{table}

\begin{table}[]
\centering
\caption{Prompt example for generating possible outlier labels based on the provided ID label space and the major category.}
\label{table:prompt-example} 
\begin{tabular}{|@{}p{0.9\linewidth}@{}|}
\hline
\textbf{Input:} \\
\textbf{Task:} \\
Generate 10 possible paper topics that belong to \textbf{Computer Science} but are distinct from the provided topics \{agents, information retrieval, database, \\
artificial intelligence\}. \\
Output : $[\{\text{"answer"}: \langle \text{your\_answer} \rangle \}, \{\text{"answer"}: \langle \text{your\_answer} \rangle \}, \dots]$.\\
\textbf{Output:} \\
\hline
\end{tabular}
\label{prompt-2}
\end{table}

\begin{table}[]
\centering
\caption{Full prompt example with Hard-Reject scenarios for zero-shot coarse classifier on the Citeseer dataset.}
\label{table:prompt-example} 
\begin{tabular}{|@{}p{1\linewidth}@{}|}
\hline
\textbf{Input:} \\
Paper: \\
Logical Case Memory Systems: Foundations And Learning Issues The focus of this paper is on the introduction of a quite general type of case-based reasoning systems called logical case memory systems.  The development of the underlying concepts has been driven by investigations in certain problems of case-based learning. Therefore, the present development of the target concepts is accompanied by an in-depth discussion of related learning problems. Logical case memory systems provide some formal framework for the investigation and for the application of structural similarity concepts. Those concepts have some crucial advantage over traditional numerical similarity concepts: The result of determining a new case\'s similarity to some formerly experienced case can be directly taken as a basis for performing case adaptation. Essentially, every logical case memory system consists of two constituents, some partially ordered case base and some partially ordered set of predicates. Cases are terms, in a logical sense. Given some problem case, every predicat... \\
\textbf{Task:} \\
There are the following categories: \\
$[\text{agents, information retrieval, database, artificial intelligence}]$. \\
If True, specify which category in \\
$[\text{agents, information retrieval, database, artificial intelligence}]$ the paper belongs to. \\
\textbf{If False, provide a suggested category that is not in the category list. The suggested category includes but not limited to the following:} [Quantum Computing and Its Applications in Cryptography, Blockchain Technology and Decentralized Applications, Edge Computing, Natural Language Processing for Sentiment Analysis, Cybersecurity, Human-Computer Interaction, Machine Learning for Predictive Maintenance in Industrial Systems, Computer Vision, Augmented Reality, Big Data Analytics, ...] \\
Provide your answer, a confidence number between [0-1], and suggested category. \\
$[ \{ \text{"answer"} : \langle \text{True or False} \rangle, \, 
   \text{"confidence"} : \langle \text{confidence\_here} \rangle, \, 
   \text{"category"} : \langle \text{category\_here} \rangle \} ]$

\textbf{Output:} \\
\hline
\end{tabular}
\label{citeseer-prompt}
\end{table}

\begin{table}[]
\centering
\caption{Full prompt example with post-OOD label space for zero-shot OOD classifier on the Citeseer dataset.}
\label{table:prompt-example} 
\begin{tabular}{|@{}p{0.9\linewidth}@{}|}
\hline
\textbf{Input:} \\
Paper: \\
Logical Case Memory Systems: Foundations And Learning Issues The focus of this paper is on the introduction of a quite general type of case-based reasoning systems called logical case memory systems.  The development of the underlying concepts has been driven by investigations in certain problems of case-based learning. Therefore, the present development of the target concepts is accompanied by an in-depth discussion of related learning problems. Logical case memory systems provide some formal framework for the investigation and for the application of structural similarity concepts. Those concepts have some crucial advantage over traditional numerical similarity concepts: The result of determining a new case\'s similarity to some formerly experienced case can be directly taken as a basis for performing case adaptation. Essentially, every logical case memory system consists of two constituents, some partially ordered case base and some partially ordered set of predicates. Cases are terms, in a logical sense. Given some problem case, every predicat... \\
\textbf{Task:} \\
There are the following categories: \\
\{machine learning, human computer interaction, \\ augmented, edge, cybersecurity, big\}. 
Which category does this paper belong to? Provide your best guess with a confidence ranging from 0 to 1. For example: \\
$[ \{ \text{"answer"} : \langle \text{category\_here} \rangle, \, 
   \text{"confidence"} : \langle \text{confidence\_here} \rangle \} ]$ \\

\textbf{Output:} \\
\hline
\end{tabular}
\label{ood-prompt}
\end{table}

\section{Proof}
\label{proof}
\subsection{Proof of Theorem 3.1}
\begin{proof}
Let the input ID space be denoted by $\mathcal{X} $, where each input $\mathbf{x} \in \mathcal{X} $ belongs to an ID class $ y_i \in \mathcal{Y}_{\text{id}} $. Let $ f(\mathbf{x}) $ denote the model's output probability distribution over the classes $ \mathcal{Y}_{\text{id}} \cup \{y_{\text{ood}}\} $, where $y_\text{ood}$ is the OOD class under a open-set setting.

To produce generated OOD samples, in the traditional manifold mixup within ID classes, the representation of the new sample \( \boldsymbol{h}_{\text{mix}} \) is created by linearly interpolating between the hidden embeddings of two different ID samples \( \boldsymbol{h}_i \) and \( \boldsymbol{h}_j \) as:
\begin{equation*}
 (\boldsymbol{h}_{\text{mix}}, y_{\text{mix}}) := (\lambda \boldsymbol{h}_i + (1-\lambda) \boldsymbol{h}_j, y_{\text{ood}}), \quad \lambda \in [0, 1] .  
\end{equation*}

Denote the generated OOD samples, which lie in the convex hull of the ID classes, as $\mathcal{X}_{\text{mix}}$. These samples encourage the model to generalize by smoothing the decision boundary between ID regions in the feature space. However, such generated samples do not explicitly extend the feature space towards new region.

Assume that LLMs possess expert knowledge to understand the categories in $\mathcal{Y}_{\text{id}}$. In the first step of CFC, the coarse-grained classifier is able to detect certain outlier data. These outlier samples related to practical test space are realistic and carry semantic meaning. Thus, CFC explore a new space different from ID space. For these semantic OOD samples, let their mean hidden representation be denoted as $\mathbf{h}_{\text{ood}}$. By applying manifold mixup between the hidden embedding of an ID sample \( \mathbf{x}_i \) and OOD center \( \mathbf{x}_{\text{ood}} \), the mixing would occur as:
\begin{equation*}
 (\boldsymbol{h}^{'}_{\text{mix}}, y^{'}_{\text{mix}}) := (\lambda \boldsymbol{h}_i + (1-\lambda) \boldsymbol{h}_{\text{ood}}, y_{\text{ood}}), \quad \lambda \in [0, 1] .  
\end{equation*}
Denote the generated OOD samples using the improved manifold mixup method as ${\mathcal{X}_{\text{mix}}^{'}}$. 



Let $\mathcal{H}$ be a vector space of dimension $\text{dim}(\mathcal{H})$. Assume the dimension of the ID input space is $\text{dim}(\mathcal{H})$. The dimension of $\mathcal{X}_{\text{mix}}$ generated within the ID space is also $\text{dim}(\mathcal{H})$. Let the dimension of the new space of semantic OOD samples be $\text{dim}(\mathcal{H}^{'})$. Then, the dimension of ${\mathcal{X}_{\text{mix}}^{'}}$ is $\text{dim}(\mathcal{H} + \mathcal{H}^{'})$, where $\text{dim}(\mathcal{H} + \mathcal{H}^{'}) > \text{dim}(\mathcal{H})$.

As the number of ID and OOD classes is $C + 1$, the representations of the training points lie fully on a $\text{dim}(\mathcal{H}) - (C + 1)$ subspace for pure ID class mixup, and on a $\text{dim}(\mathcal{H} + \mathcal{H}^{'}) - (C + 1)$ subspace for our ID and OOD mixed augmentation.

We have $\text{dim}(\mathcal{H} + \mathcal{H}^{'}) - (C + 1) > \text{dim}(\mathcal{H}) - (C + 1)$. Thus, compared to manifold mixup applied only to ID classes, the mixup between ID and OOD classes extends the OOD feature space beyond the convex hull of ID samples. Consequently, CFC results in a smoother and flatter decision boundary for OOD detection.

\end{proof}

\subsection{Manifold Mixup-based Subspace Extension on Classifier Decision Boundary}

We provide a theoretical analysis of how embedding mixup expands the representation subspace and influences the classifier’s decision boundary.

Let $\mathcal{X} = {\mathbf{x}i}{i=1}^n \subset \mathbb{R}^d$ denote the graph node features, and $\mathcal{Y} = {y_i}_{i=1}^n$ their corresponding labels, where $y_i \in {1, \dots, C}$. We consider a simple linear classifier $f(\mathbf{x}) = \text{softmax}(W^\top \mathbf{x})$ trained on labeled data.

To regularize the classifier and expand the representation space, we apply manifold mixup in the latent space. Given two embeddings $\mathbf{x}_i$ and $\mathbf{x}_j$, the mixed feature and its corresponding soft label are given by:
\begin{equation}
\begin{aligned}
\small
& \tilde{\mathbf{x}} = \lambda \mathbf{x}_i + (1 - \lambda)\mathbf{x}_j, \quad \lambda \sim \text{Beta}(\alpha, \alpha), \\
& \tilde{\mathbf{y}} = \lambda \mathbf{y}_i + (1 - \lambda)\mathbf{y}_j.
\end{aligned}
\end{equation}
In CFC, we perform OOD data augmentation using manifold mixup on the latent representations of identified OOD samples. Let $\mathcal{H}'$ denote the subspace spanned by the OOD embeddings. The mixup operation constructs an extended subspace $\widetilde{\mathcal{H}'} \supset \mathcal{H}'$.

As shown in \cite{verma2019manifold}, the mixup-based training objective:

\begin{equation}
\mathcal{L}{\text{mix}} = \mathbb{E}{\lambda, i, j} \left[ \mathcal{L}(f(\tilde{\mathbf{x}}), \tilde{\mathbf{y}}) \right]
\end{equation}

can be interpreted as introducing an implicit regularization effect:

\begin{equation}
\mathcal{L}{\text{mix}} \approx \mathcal{L}{\text{emp}} + \Omega(W),
\end{equation}

where $\mathcal{L}_{\text{emp}}$ is the empirical risk, and $\Omega(W)$ penalizes abrupt changes in the decision function. This regularization encourages smoother transitions between classes, effectively enlarging inter-class margins and promoting a more robust, well-structured decision boundary for both ID and OOD detection.

\section{Datasets}
\label{datasets}
\begin{table*}[t]
\centering
\caption{The details of graph datasets.}
\renewcommand\arraystretch{1.2}

\resizebox{0.95\textwidth}{!}{%
\begin{tabular}{p{2.5cm} | c c c c p{7.5cm}}
\hline
\textbf{Dataset} & \textbf{Nodes} & \textbf{Edges} & \textbf{Features} & \textbf{Labels} & \textbf{Categories} \\
\hline

Cora & 2708 & 5429 & 1433 & 7 &
Rule Learning, Neural Networks, Case Based, Genetic Algorithms, Theory, Reinforcement Learning, Probabilistic Methods \\
\hline

Citeseer & 3312 & 4732 & 3703 & 6 &
Agents, Machine Learning, Information Retrieval, Database, Human Computer Interaction, Artificial Intelligence \\
\hline

WikiCS & 11,701 & 216,123 & 300 & 10 &
Computational Linguistics, Databases, Operating Systems, Computer Architecture, Computer Security, Internet Protocols, Computer File Systems, Distributed Computing Architecture, Web Technology, Programming Language Topics \\
\hline

DBLP & 17,716 & 105,734 & 1639 & 4 &
Data Mining, Database, Artificial Intelligence, Information Retrieval \\
\hline

Amazon-Computer & 87,229 & 1,256,548 & -- & 10 &
Laptop Accessories, Computer Accessories and Peripherals, Computer Components, Storage, Networking Products, Monitors, Computers and Tablets, Tablet Accessories, Servers, Tablet Replacement Parts \\
\hline

Amazon-Photo & 48,362 & 873,782 & -- & 12 &
Film Photography, Video, Digital Cameras, Accessories, Binoculars and Lenses, Bags and Cases, Lighting and Studio, Flashes, Tripods and Monopods, Underwater Photography, Video Surveillance \\
\hline

\end{tabular}
}

\label{data}
\end{table*}

\begin{table*}[]
    \centering
    \caption{The details of text datasets.}
    \renewcommand\arraystretch{1.2}
    \begin{center}
     \resizebox{0.95\textwidth}{!}{%
    \begin{tabular}{p{2.5cm} | p{1cm} c c}
        \hline
        \textbf{Dataset} & \textbf{Number} & \textbf{Labels} & \textbf{Categories} \\
        \hline
       \multirow{3}{*}{News Category} &  \multirow{3}{*}{210, 000}& \multirow{3}{*}{17} & arts, business, crime, education, entertainment, environment, \\
       & & & food \& drink, healthy living, home \& living, news, parents, \\
       & & & politics, queer voices, religion, science, style, travel \\
       \hline
        \multirow{2}{*}{Twitter Topic} &  \multirow{2}{*}{3184} & \multirow{2}{*}{6} & arts \& culture, business \& entrepreneurs, daily life \\
        & & & pop culture, science \& technology, sports \& gaming \\
        \hline
        \multirow{4}{*}{20 Newsgroups} &  \multirow{4}{*}{18, 000} & \multirow{4}{*}{20} & {alt.atheism, comp.graphics, comp.os.ms-windows.misc, comp.sys.ibm.pc.hardware, } \\
        
        & & & {comp.sys.mac.hardware, comp.windows.x, misc.forsale, rec.autos, rec.motorcycles, } \\
        & & & {rec.sport.baseball, rec.sport.hockey, sci.crypt, sci.electronics, sci.med, sci.space, } \\
        & & & {soc.religion.christian, talk.politics.guns, talk.politics.mideast, talk.politics.misc, talk.religion.misc} \\
        \hline
    \end{tabular}
    }
    \end{center}
    \label{data}
\end{table*}

\section{Other Experimental Settings and Results for Graph Node Classification}
\label{graph-task}

\subsection{Details of Baselines}
\textbf{CCN\_softmax}: GCN (\cite{gcn}) with a softmax layer is used as the final output layer.

\textbf{CCN\_sigmoid}: The softmax layer of GCN is replaced with multiple 1-vs-rest of sigmoids. 

\textbf{CCN\_softmax\_{$\tau$}}: Based on GCN\_softmax, a probability threshold chosen in $\{0.1, 0.2, . . . , 0.9\}$ is used for the classification of each class.

\textbf{CCN\_sigmoid\_{$\tau$}}: Based on GCN\_sigmoid, a probability threshold chosen in $\{0.1, 0.2, . . . , 0.9\}$ is used for the classification of each class.

\textbf{GCN\_Poser}: Employing the method in Poser (\cite{zhou2021learning}) with GCN.

$\mathbf{{G}^{2}Pxy}$: Using two kinds of proxy unknown nodes for unknown node detection.

\textbf{GNNSafe}: proposing an effective OOD discriminator based on an energy function extracted from graph neural networks.

For fair comparisons, CFC employs GCN as the backbone for GNN-based fine classification, consistent with the other methods. All experiments are repeated five times using different random seeds.

\subsection{Settings for OOD classes}

In the case of two OOD classes ($u$=2), the classes Rule Learning and Case Based are selected as OOD for Cora, Machine Learning and Human-Computer Interaction for Citeseer, Web Technology and Programming Language Topics for WikiCS, and Artificial Intelligence and Information Retrieval for DBLP. 

In the case of three OOD classes ($u=3$), the classes Rule Learning, Case Based, and Probabilistic Methods are selected as OOD for Cora, Machine Learning, Human-Computer Interaction, and DataBase for Citeseer, Distributed computing architecture, Web Technology and Programming Language Topics for WikiCS. 

\subsection{Hyperparameter analysis}
\label{app-parameter}
We analyze the impact of the parameter $\alpha$, which balances ID and OOD features in Equation $\tilde{x}_i=\alpha \boldsymbol{h}_i^k + (1-\alpha) \boldsymbol{h}_{c}^k, i \leq K$, where $\boldsymbol{h}_i^k$ is the hidden embedding of ID nodes and $\boldsymbol{h}_c^k$ denotes the center embedding of identified OOD samples from the coarse classifier with LLMs. The results in the following table show that although the performance of CFC varies with $\alpha$, it consistently outperforms other OOD detection methods.

\begin{table*}[]
	\caption{The results of CFC with different values of parameter $\alpha$.}
	\centering
	\renewcommand\arraystretch{1.1}
	\begin{center}
            \resizebox{0.95\textwidth}{!}{%
		\begin{tabular}{l|cc|cc|cc|cc|cl}
			\toprule[1pt]
			\multirow{2}{*}{Dataset} & \multicolumn{2}{c|}{0.1} & \multicolumn{2}{c|}{0.3} & \multicolumn{2}{c|}{0.5} & \multicolumn{2}{c|}{0.7} & \multicolumn{2}{c}{0.9}\\
			\cline{2-11}
    & ID & OOD & ID & OOD & ID & OOD & ID & OOD & ID & OOD \\
    \hline
    Cora & 90.04$\pm$0.43 & 82.13$\pm$0.81 & 89.92$\pm$0.34 & 83.09$\pm$0.93 & 86.01 $\pm$0.96 & 96.21$\pm$0.61 & 86.02$\pm$0.96 & 96.22$\pm$0.61 & 86.02$\pm$0.96 & 96.22$\pm$0.61\\
			\bottomrule[1pt]
		\end{tabular}
  }
	\end{center}
	\label{tab-noisy}	
\end{table*}

\subsection{Robustness of CFC under limited ID data}
\label{robust}
We evaluated the robustness of CFC under limited ID data. As shown in Table \ref{tab-robust}, reducing ID training data may slightly decrease ID classification accuracy, but can in some cases improve OOD detection.

\begin{table}[H]
	\caption{The results on Cora and Citeseer. Here 0.2 means we use 20\% training data compared to the ID training data in our paper.}
	\centering
	\renewcommand\arraystretch{1.1}
	\begin{center}
            \resizebox{0.5\textwidth}{!}{%
		\begin{tabular}{l|ccc|ccl}
		\hline
			\multirow{2}{*}{Ratio} & \multicolumn{3}{c|}{Cora} & \multicolumn{3}{c}{Citeseer} \\
						 \cline{2-7}    
            & ID &  OOD & overall  & ID & OOD & overall \\
            \hline
             0.2 & 81.28$\pm$2.12 & 93.73$\pm$1.47 & 85.07$\pm$1.20 & 69.98$\pm$1.84 & 86.07$\pm$1.49 & 77.87$\pm$0.37\\
             0.5 & 84.09$\pm$1.67 & 95.12$\pm$0.91 & 87.45$\pm$0.10 & 70.82$\pm$0.96 & 84.96$\pm$0.90 & 77.82$\pm$0.49\\
             1 & 87.49$\pm$0.96 & 95.74$\pm$0.61 & 90.00$\pm$0.54 & 73.92$\pm$1.02 & 80.57$\pm$0.78 & 77.21$\pm$0.82 \\
			\hline
		
		\end{tabular}
  }
	\end{center}
	\label{tab-robust}	
\end{table}

\subsection{The results of various LLMs as feature encoder}
\label{app-encoders}

Table \ref{main-encoder} compares different LLM feature encoders. Overall, e5-large-v2 performs better and more stably in joint training. Additionally, LLMs specialize in different topics—for instance, Llama2 excels in ID topics for Citeseer, while e5 and ST perform well in OOD.

\begin{table*}[]
    \centering
    \caption{Comparison of various LLMs as feature encoder on ID classification and OOD detection across four datasets.}
    \begin{center}
    \renewcommand\arraystretch{1.2}
\resizebox{0.95\textwidth}{!}{%
    \begin{tabular}{l|ccc|ccc|ccc|ccc}
        \hline
        \multirow{2}{*}{LLM Encoder} & \multicolumn{3}{c|}{Cora} & \multicolumn{3}{c|}{Citeseer} & \multicolumn{3}{c|}{WikiCS} & \multicolumn{3}{c}{DBLP}\\
        \cline{2-13}
                                 & ID &  OOD & overall  & ID & OOD & overall & ID & OOD & overall & ID & OOD & overall\\
        \hline
        CFC-e5 & 87.49 &  95.74 & 90.00 & 73.92 & 80.57 & 77.21 &  81.22 & 75.93 & 79.73 & 76.06 & 73.61 & 74.44 \\
        CFC-ST & 84.24 & 90.03 & 86.00 & 72.78 & 81.71 & 77.20 & 78.84 & 65.35 & 75.02 & 83.02 & 70.61 & 74.82\\ 
        CFC-Llama2-7b & 84.72 & 92.99 & 87.24 & 76.06 & 73.14 & 74.62 & 79.05 & 73.10 & 77.37 & 83.04 & 50.30 & 61.42\\  
        CFC-Llama2-13b & 82.14 & 94.98 & 86.05 & 76.53 & 72.29 & 74.43 & 75.55 & 68.06 & 73.43 & 87.14 & 45.42 & 59.59\\
        \hline
    \end{tabular}
    }
    \end{center}
    \label{main-encoder}
    \vskip -0.1in
\end{table*}

\subsection{Other Results of graph tasks}

With an increasing number of OOD samples in the test set, we observe higher OOD accuracies for the case of three OOD classes in Table \ref{main-ood3} in compared to two OOD classes in Table \ref{main-ood2} for the proposed CFC on Citeseer and WikiCS.

\begin{table}[]
    \centering
    \caption{Comparison of various methods on ID classification and OOD detection across three datasets: Cora, Citeseer, WikiCS (3 OOD classes).}
    \begin{center}
    \renewcommand\arraystretch{1.2}
\resizebox{0.5\textwidth}{!}{%
    \begin{tabular}{l|ccc|ccc|ccc}
        \hline
        \multirow{2}{*}{Methods} & \multicolumn{3}{c|}{Cora} & \multicolumn{3}{c|}{Citeseer} & \multicolumn{3}{c}{WikiCS} \\
        \cline{2-10}
                                 & ID &  OOD & overall  & ID & OOD & overall & ID & OOD & overall \\
        \hline
        GCN\_softmax         &  91.53 & 0.00 & 45.61 & 74.56 & 0.00 & 22.55 & 61.42 & 0.00 & 38.03  \\
        GCN\_sigmoid    & 92.42 & 0.00 & 46.05 & 79.21 & 0.00 & 23.96 & 70.62 & 0.00 & 43.73 \\        
        GCN\_softmax\_$\tau$   & 61.37 & 74.13 & 67.69 & 35.62 & 91.50 & 74.61 & 35.23 & 70.82 & 48.79  \\
        
        GCN\_sigmoid\_$\tau$    &  77.10 & 57.16 & 66.85 & 45.90 & 88.22 & 75.38 & 45.15 & 86.90 & 61.04 \\
        GCN\_PROSER    & 71.01 & 80.31 & 75.61 & 53.96 & 77.09 & 70.10    & 50.94 & 90.20 & 65.49       \\
        ${G}^2Pxy$ & 69.62 & 76.62  &  73.21 & 61.43 & 76.92 & 72.23  & 49.57 & 92.07 & 64.88 \\
        \hline
        CFC (wo / D/M) & 85.08 & 86.67 & 85.89 & 63.55 & 91.75 & 83.64 & 76.96 & 88.84 & 81.53 \\
        CFC (wo / M) & 89.98 & 81.20 & 85.52 & 67.57 & 90.11 & 83.63 & 81.75 & 75.39 & 79.30 \\
        CFC (wo / D) & 82.48  & 89.04 & 85.81 & 63.97 & 92.17 & 84.06  & 76.63 & 90.13 & 81.82 \\  
        CFC & 83.23 & 91.87 & 87.62 & 66.73 & 91.40 & 84.30 & 80.66 & 81.64 & 81.04\\
        \hline
    \end{tabular}
    }
    \end{center}
    \label{main-ood3}
\end{table}

\begin{table}[] 
    \centering
    \caption{The results of OOD classification with different LLM encoder.}
    \resizebox{0.5\textwidth}{!}{%
        \begin{tabular}{l|p{1.2cm} p{1.2cm}p{1.2cm}c}
            \hline
            {Encoder} & {Cora} & {Citeseer} & {WikiCS} & {DBLP} \\
            \hline
            GPT-4o & 67.76 & 70.30 & 57.96 & 48.45\\
            Llama2 & 59.45 & 65.88 & 15.40 & 59.97\\
            Llama3 & 62.20 & 69.83 & 15.00 & 50.73 \\
            \hline
        \end{tabular}
    }
    \label{llm-ood}
    \vskip -0.1in
\end{table}

\section{Evaluation on Other Task}
\label{text-task}
We show that the proposed CFC is a general framework for comprehensively solving open-set classification problem including ID classification and OOD classification by applying CFC on text domain.

\subsection{Introduction of Datasets}
We use two text datasets including News Category dataset \cite{misra2022news}, and Twitter Topic dataset \cite{antypas2022twitter}.

\paragraph{News Category Dataset} is one of the largest news datasets, containing approximately 210k news headlines from HuffPost, published between 2012 and 2022. The dataset includes 42 classes, which are significantly imbalanced. To mitigate confusion among similar classes, closely related categories were merged. As a result, the dataset was reduced to 17 distinct representative classes as in \cite{baran2023classical}. We use 7 classes—entertainment, healthy living, news, parents, politics, style, and travel—as the ID space, with the remaining 10 classes designated as OOD. The ID data is split for training, validation, and testing following the same procedure as \cite{baran2023classical}. Additionally, 30\% of the OOD data is selected for validation, with the remaining portion reserved for testing.


\paragraph{Twitter Topic Classification} is a topic classification dataset derived from Twitter posts. It contains 3,184 high-quality tweets, each categorized into one of six distinct classes.

\subsection{Baselines}

\paragraph{Maximum Softmax Probability(MSP) } employs the softmax score to check the certainty of whether a sample belongs to a domain.

\paragraph{Energy-based} \cite{liu2020energy} leverages an energy score function to quantify the model's confidence.

\paragraph{Rectified Activations (ReAct)} \cite{sun2021react} is a straightforward method for mitigating model overconfidence on OOD examples by truncating excessively high activations during evaluation.

\paragraph{KL-Matching (KLM)} \cite{hendrycks2019scaling} computes the minimum KL-divergence between the softmax probabilities and the mean class-conditional distributions.

\paragraph{GradNorm} \cite{huang2021importance} uses the vector norm of the gradients to differentiate between in-distribution (ID) and out-of-distribution (OOD) samples, operating under the assumption that larger norm values are indicative of in-distribution data.

\paragraph{Directed Sparsification (DICE)} \cite{sun2022dice} employs selective sparsification to retain a subset of weights, effectively removing irrelevant information from the model's output.

\paragraph{Virtual-logit Matching (ViM)} \cite{wang2022vim} integrates information from both the feature space and the output logits. This approach provides a combination of class-agnostic and class-dependent knowledge, enabling improved separation of OOD data.

\paragraph{K-nearest Neighbors (KNN)} \cite{sun2022out} calculates the distance between the embedding of a given input and the embeddings from the training set to determine whether the input belongs to the in-distribution (ID) or not.

\subsection{Experiment Setting}
For all methods, we use the RoBERTa\textsubscript{base} model (Liu et al., 2019) as the backbone, paired with a fully connected layer for classification. All baseline methods are fine-tuned on the training data with a learning rate of 0.0001, using the Adam optimizer with a weight decay of $5\times10^{-4}$.

In the proposed CFC approach, OOD samples with low confidence scores from the LLM (below 0.8) are removed during the denoising procedure. For data augmentation, the identified OOD samples from the LLM are mixed with other samples in a batch to generate additional synthetic OOD examples. To ensure the robustness of the results, all experiments are repeated five times with different initial seeds to minimize the impact of randomness.


\begin{table}[]
    \centering
    \caption{Comparison of various methods on OOD detection across two datasets: News Category and Twitter.}
    \begin{center}
    \renewcommand\arraystretch{1.2}
    \resizebox{0.4\textwidth}{!}{%
    \begin{tabular}{l|c|l}
        \hline
        {Methods} & {News Category} & {Twitter} \\
        \hline
        msp             & 71.07  & 61.83  \\
        Energy-based    & 71.89  & 53.52  \\        
        ReAct           & 72.16  & 56.15  \\
        KLM             & 60.12 & 50.03   \\
        GradNorm        & 71.33 & 51.54   \\
        DICE            & 64.32  & 44.68   \\
        Vim             & 75.33 & 60.89   \\
        KNN             & 76.20  & 58.66  \\
        \hline
        CFC             & 82.04 & 71.68  \\
        \hline
    \end{tabular}
    }
    \end{center}
    \label{main-text}
\end{table}

\begin{table}[]
    \centering
    \caption{The accuracy of single OOD category across two datasets: News Category and Twitter.}
    \begin{center}
    \renewcommand\arraystretch{1.2}
    \resizebox{0.5\textwidth}{!}{%
    \begin{tabular}{c|c|c|c|c|c}
        \hline
        \multicolumn{3}{c|}{News Category} & \multicolumn{3}{c}{Twitter} \\
        \hline
        ID & OOD & Accuracy & ID & OOD & Accuracy \\
        \hline
        entertainment & arts & 50.48 &  & &  \\        
        healthy\_living & crime & 66.32 &  &  &  \\
        news & education & 46.43 & arts\_\&\_culture & pop\_culture & 37.45 \\
        parents & home\_\&\_living & 16.67 & business\_\&\_entrepreneurs & science\_\&\_technology & 32.08 \\
        politics & queer\_voices & 27.90 & daily\_life & sports\_\&\_gaming & 39.70 \\
        style & science & 30.00 &  &  & \\
        travel & & &  &  & \\
        \hline
    \end{tabular}
    }
    \end{center}
    \label{ood-text}
\end{table}

\subsection{Results}

Table \ref{main-text} presents the results of CFC compared to other OOD detection methods on text datasets. CFC consistently outperforms all baseline methods, mirroring the improvements observed in graph datasets.

To further validate CFC's effectiveness in OOD classification for text data, we report the results in Table \ref{ood-text}. In the News Category dataset, 7 classes are used as ID labels, while the Twitter Topic dataset has an ID label space of 3 classes. Our findings demonstrate that CFC successfully identifies various OOD classes, highlighting its capability in addressing the comprehensive open-world classification problem.

\end{document}